\documentclass[conference]{IEEEtran}
\IEEEoverridecommandlockouts
\usepackage{cite}
\usepackage{amsmath,amssymb,amsfonts}
\usepackage{algorithmic}
\usepackage{graphicx}
\usepackage{textcomp}
\usepackage{xcolor}
\def\BibTeX{{\rm B\kern-.05em{\sc i\kern-.025em b}\kern-.08em
    T\kern-.1667em\lower.7ex\hbox{E}\kern-.125emX}}

\usepackage{amsfonts}
\usepackage{todonotes}
\usepackage{lipsum}
\usepackage{wrapfig}
\usetikzlibrary{calc,arrows,arrows.meta,backgrounds,tikzmark,shapes.geometric, decorations.pathreplacing,decorations.markings, spy}
\usepackage{colortbl} %needed for the tables form nerf
\usepackage{booktabs}
\usepackage{array,multirow}
\usepackage[binary-units=true]{siunitx}
\usepackage{bm}
\usepackage{multirow, tabularx}
\newcolumntype{C}{>{\centering\arraybackslash}X}
\PassOptionsToPackage{printonlyused,smaller,nolist}{acronym}
\usepackage{acronym} % nice macros for handling all acronyms
\usepackage{xspace}
\usepackage{url}
\def\etal{\emph{ et al}.}
\newcommand{\reffig}[1]{Fig.~\ref{#1}}
\newcommand{\reftab}[1]{Tab.~\ref{#1}}

\DeclareSIUnit{\nothing}{\relax}

\definecolor{ph-purple}{RGB}{129, 39, 232}
\definecolor{ph-blue}{RGB}{5, 131, 227}
\definecolor{ph-gray}{rgb}{0.5, 0.5, 0.5}
\definecolor{ph-orange}{RGB}{227, 127, 5}
\definecolor{ph-green}{RGB}{0, 135, 124}
\definecolor{ph-yellow}{RGB}{235, 201, 52}
\definecolor{ph-light-green}{RGB}{181, 209, 21}
\definecolor{ph-red}{RGB}{250, 101, 60}
\colorlet{ph-orange-light}{ph-orange!70}
\colorlet{ph-blue-light}{ph-blue!70}
\colorlet{ph-purple-light}{ph-purple!70}
\colorlet{ph-green-light}{ph-green!70}
\definecolor{ph-light-gray}{rgb}{0.75, 0.75, 0.75}

\newcommand{\norm}[1]{\left\lVert#1\right\rVert}
\begin{document}
	
% required after document
% using nolist to not produce the acro list
\begin{acronym}
	\acro{nvs}[NVS]{Novel View Synthesis}
	\acro{mvs}[MVS]{Multi-View Stereo}
	\acro{nerf}[NeRF]{Neural Radiance Field}
	\acro{srn}[SRN]{Scene Representation Networks}
\end{acronym}

\title{NeuralMVS: Bridging Multi-View Stereo and Novel View Synthesis\\
}

%\author{\IEEEauthorblockN{Anonymous Authors}}

\author{\IEEEauthorblockN{Radu Alexandru Rosu}
	\IEEEauthorblockA{\textit{Autonomous Intelligent Systems} \\
		\textit{University of Bonn}\\
		Bonn, Germany \\
		rosu@ais.uni-bonn.de}
	\and
	\IEEEauthorblockN{Sven Behnke}
	\IEEEauthorblockA{\textit{Autonomous Intelligent Systems} \\
		\textit{University of Bonn}\\
		Bonn, Germany \\
		behnke@ais.uni-bonn.de}
	
	\thanks{This work has been funded by the Deutsche Forschungsgemeinschaft (DFG, German Research Foundation) under Germany's Excellence Strategy - EXC 2070 - 390732324.}%
}

%\and
%\and
%\and
%\and
%\and

\maketitle

\begin{abstract}
\ac{mvs} is a core task in 3D computer vision. With the surge of novel deep learning methods, learned \ac{mvs} achieves more complete depth maps than classical approaches, but still relies on building a memory intensive dense cost volume. \ac{nvs} is a parallel line of research and has recently seen an increase in popularity with \ac{nerf} models, which optimize a per scene radiance field. However, \ac{nerf} methods do not generalize to novel scenes and are slow to train and test.  
We propose to bridge the gap between these two methodologies with a novel network that can recover 3D scene geometry as a distance function, together with high-resolution color images. Our method uses only a sparse set of images as input and can generalize well to novel scenes. Additionally, we propose a coarse-to-fine sphere tracing approach in order to significantly increase speed.
We show on various datasets that our method reaches comparable accuracy to per-scene optimized methods while being able to generalize and running significantly faster. We provide the source code at \normalfont{\url{https://github.com/AIS-Bonn/neural_mvs}}
\end{abstract}

%\begin{IEEEkeywords}
%image-based, novel-view, multi-view
%\end{IEEEkeywords}

\section{Introduction}
%This document is a model and instructions for \LaTeX.
%Please observe the conference page limits. 

Multi-view Stereo (MVS) recovers depth and geometry from multiple images with known camera poses. This is usually done with classical methods like COLMAP~\cite{colmap}, Gipuma~\cite{gipuma}, or MVE~\cite{mve} by searching for correspondences along epipolar lines. These algorithms lack learned components and cannot cope with challenging conditions like imperfect calibration, blurry or incomplete images, and heavy occlusion~\cite{laga2019survey, yao2018mvsnet}. 

Recent Novel View Synthesis (NVS) methods like NeRF~\cite{nerf} recover geometry as a byproduct of view synthesis. Geometry is represented as a radiance field which is multi-view consistent between all images. This has the advantage of recovering more complete depth while being more robust to imperfect images than classical methods. Main disadvantages of NVS methods are the large processing time and a lack of generalization. They require per-scene training which can take up to several days.
Additionally, inference speed is also limited---often requiring multiple minutes to synthesize a full image together with the corresponding depth.

More recent \ac{nvs} approaches like pixelNeRF~\cite{pixelnerf} and IBRNet~\cite{ibrnet} improve the reconstruction speed by having a training phase which allows the model to generalize to novel scenes and thus require only little per-scene fine-tuning. 
Inference is still slow as the network needs to query the radiance field multiple times during synthesis and requires a very dense sampling of the view frustum.

\bgroup
\def\Img{./images/teaser/teaser.png} 
\newlength{\WT}
\newlength{\HT}
\settowidth{\WT}{\includegraphics{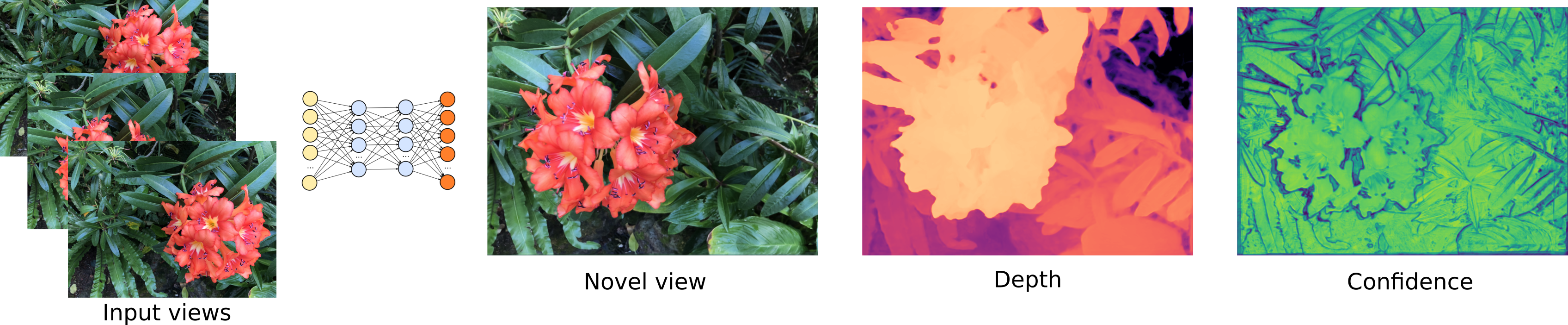}}
\settoheight{\HT}{\includegraphics{\Img}}
\begin{figure*}
	\begin{center}
		\includegraphics[trim=.0\WT{} 0.1\HT{} 0.02\WT{} 0.0\HT{},clip, width=\textwidth]{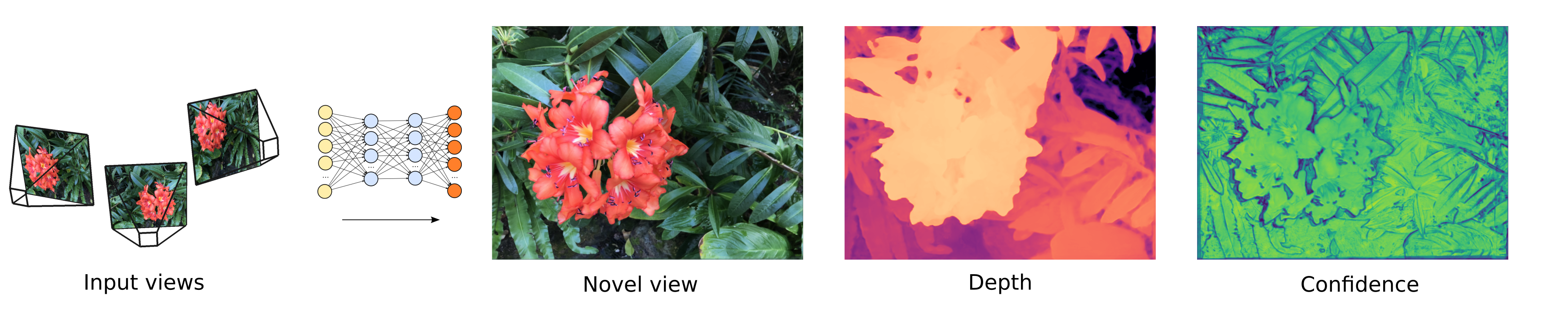}
		\vspace*{-6ex}
	\end{center}
	\caption{NeuralMVS processes multiple input views to synthesize a novel colored view with corresponding depth and gives an estimate for the output confidence.}
	\label{fig:teaser}
\end{figure*}
\egroup

In our approach, we leverage ideas from MVS and NVS and combine them in a new learning-based method that generalizes well to novel scenes and reaches comparable reconstructions to per-scene optimized methods while requiring only a fraction of the time.

First, our method accelerates the ray marching step by differentiable sphere tracing. While ray marching requires hundreds of samples per ray in order to achieve good accuracy, sphere tracing can reach the object surface with only a few iterations by predicting for each ray sample its jump towards the next one. 

Second, instead of tracing one ray per pixel, our approach starts by tracing on a coarse image which is iteratively refined until we reach the full resolution. This coarse-to-fine method further alleviates the labor-intensive step of finding the 3D surface of the object. 

Third, we propose a new scheme for the selection of conditioning views based on a Delaunay triangulation of the input views. We find that this is more temporally stable than methods based on proximity of viewing direction.

Fourth, we introduce a loss that encourages the network to output a confidence map for the novel RGB-D view. These confidence values align well with parts of the image that are undersampled or occluded and may be used to inform further reconstruction or refinement methods.

In summary, our contributions are:
\begin{itemize}
	\item a new learning-based novel view synthesis method which generalizes to unseen scenes,
	\item an efficient coarse-to-fine approach based on differentiable sphere tracing to recover depth with few samples conditioned on a set of input views, and
	\item a loss that encourages the network to output a confidence map for each novel view produced.
\end{itemize}

The general pipeline of our method is shown in~\reffig{fig:teaser} and the core components are detailed in~\reffig{fig:overview}.

\section{Related Work}
	
Several methods for learned MVS have gained popularity lately. 
MVSNet~\cite{yao2018mvsnet} proposes an end-to-end differentiable model to learn depth inference from unstructured stereo. Features are extracted from images and a dense cost volume is built using samples at regular intervals. The volume is regularized using 3D convolutions and then used to regress a depth map. In contrast in our approach, we do not define the depth samples a priori, but rather let the network learn where to sample using a differentiable sphere tracer.

The work of Darmon\etal~\cite{darmon2021deep} further builds on MVSNet and shows that depth can be recovered by using a color reconstruction loss. Similarly, we only employ an RGB loss, and do not supervise the depth map as in many settings an accurate ground-truth may not be available.

Recently, NeRF~\cite{nerf} has gained popularity for synthesizing highly-detailed novel views. NeRFs represent the scene as a radiance field and optimize it using differentiable volume rendering. The volume rendering step is computationally expensive since it samples the 3D space densely at regular intervals. One main contribution of NeRF is the introduction of the positional encoding in the context of \ac{nvs} that enables the model to learn high-frequency details.  In order to recover the 3D surface precisely, Mildenhall et al. propose a hierarchical sampling strategy that optimizes two NeRF models: one for coarse samples and one for fine samples closer to the surface. 
In our work, we leverage the positional encoding for our ray marching step and propose sphere tracing as a method to alleviate the regular sampling of NeRF.

Scene Representation Networks (SRN)~\cite{srn} is another approach which uses sphere tracing for traversing the rays in 3D space. However, this method is only able to recover low-frequency detail of the scene while we recover fine details by directly conditioning our model on the image features.

Other methods like FaDIV-Syn~\cite{rochow2021fadivsyn} propose to recover novel views of the scene without reconstructing depth by warping input views into the target frame at a series of predefined depth planes and letting the network learn how to best render the novel view. In contrast, we infer both the depth and the novel view jointly by explicitly letting the network modify the depth planes which are used to project input views.

%%%More
	MVSNerf~\cite{mvsnerf} is a general network that can recover radiance fields conditioned on input views. They explicitly construct a cost volume using sweeping planes. The cost volume is processed by 3D convolutions into a neural encoding volume and the local voxels features are used to output density and color along the ray. A limitation of their work is that the neural encoding volume is represented in the frustum of a reference view. As a result, only the contents of the scene that are visible from the reference view can be fine-tuned and rendered in high detail. In contrast, our approach doesn't define a fixed cost volume and rather aggregates image features from nearby view onto the casted rays. This allows us to model arbitrary scenes while dynamically changing the input views.

	PixelNeRF~\cite{pixelnerf} achieves generalization by aggregating features from nearby images onto the ray samples. The aggregated image features together with positional encodings are passed to a NeRF network that output final radiance and color. However, the ray sampling strategy is the same as the original NeRF, requiring hundreds of samples and slowing down inference. In contrast, we let the network learn the spacing between samples, greatly reducing their number and achieving higher rendering speed.

	Similarly to PixelNeRF, IBRNet~\cite{ibrnet} proposes to aggregate image features onto the ray. Additionally, they use a ray transformer that enables ray samples to attend to each other and better reason about occlusions. In our method, the samples along the ray can communicate front-to-back through the usage of an LSTM that dynamically predicts the jump between samples.

	Local Light Field Fusion~\cite{llff} recover novel-view by promoting input views to a local light field representation and blending them at the target view location. However, their method is only demonstrated on front facing scenes due to the multi-plane approach. Our method recovers both depth and color and can render views from arbitrary scenes.

\section{Method}

Given a set of source views, our method synthesizes depth and color image for a novel target view pose. The core idea is to recover a depth map for the target view such that warping source views onto it results in an image that matches the target as close as possible.
Our method can thus be viewed as two jointly trained networks where the first one recovers the geometry of the scene and a second network which uses that geometry as a proxy onto which the source views are projected to recover the novel view.

	\bgroup
	\def\Img{./images/overview/overview4.png} 
	\newlength{\WSA}
	\newlength{\HSA}
	\settowidth{\WSA}{\includegraphics{\Img}}
	\settoheight{\HSA}{\includegraphics{\Img}}
	\begin{figure*}
		\begin{center}
			\includegraphics[trim=.15\WSA{} 0.30\HSA{} 0.0\WSA{} 0.27\HSA{},clip, width=\textwidth]{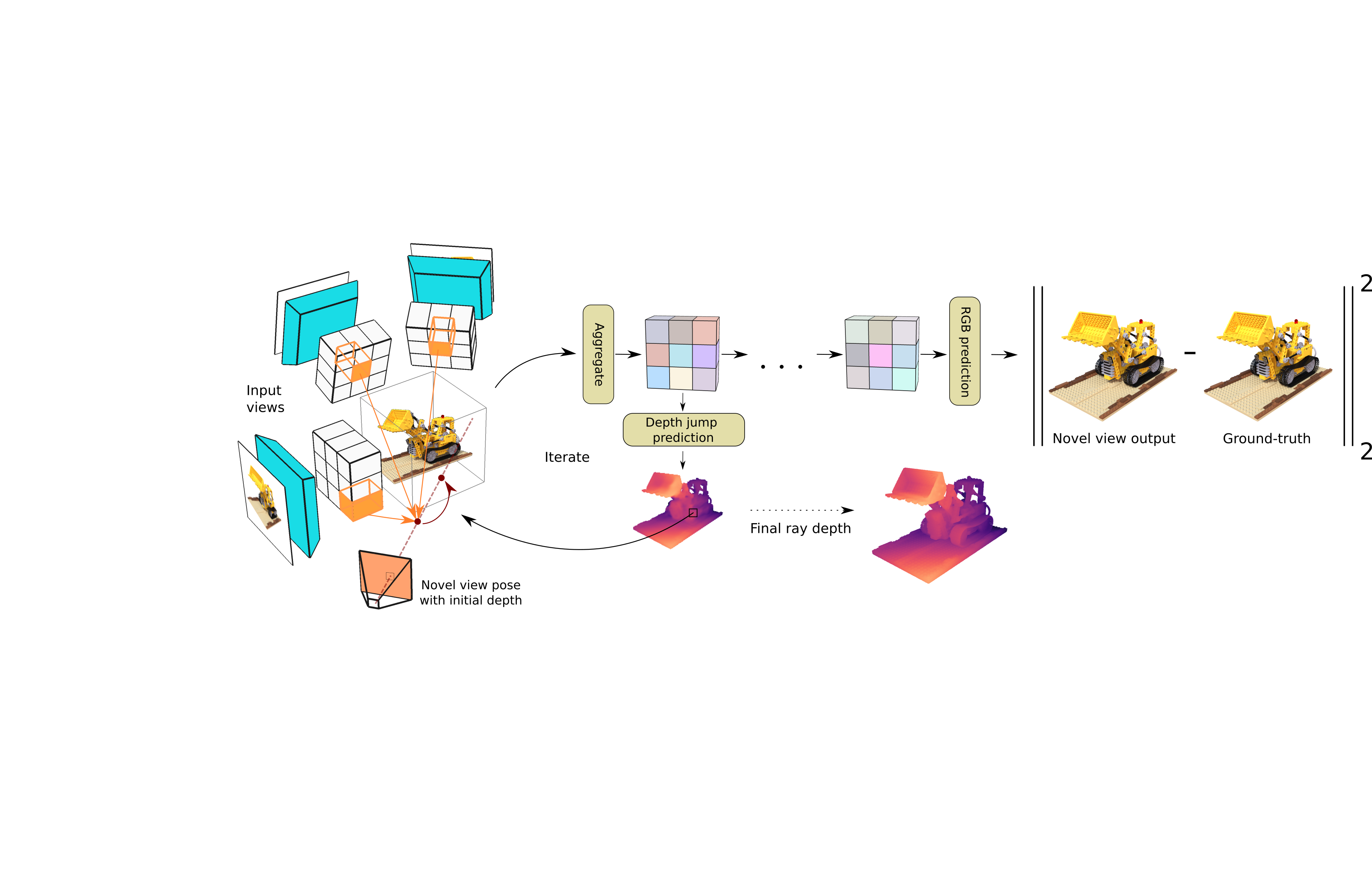}
			\vspace*{-5ex}
		\end{center}
		\caption{High-level features from the input view images are aggregated onto each ray sample. For each ray, a recursive network predicts the jump towards the next sample (red arrow). This process is repeated for a fixed number of iterations. At the last ray iteration, the final aggregated features are passed through a rendering network in order to predict the novel view RGB map. The network is only supervised with an RGB loss.}
		%	   \caption{The input views are first passes through an encoder in order to recover high-level features. From the novel views rays are shot into the scene. The depth along the ray is set initially to be close to the camera. From each ray sample, we aggregate the features from the input views in order to obtain a feature representation aligned with the novel view. We use these features to predict a jump along the ray direction which will define our new sampling position. This process is iterated several times until the ray hits the surface of the object. The final view-aligned features are then used to predict a RGB map and the processed is trained end-to-end.}
		\label{fig:overview}
	\end{figure*}
	\egroup
	
	\subsection{ View Selection Strategy  }
	In order to select the best suited source views to create the target view, various schemes have been proposed. Most of them are based on spatial proximity and view direction~\cite{colmap,mve}. However, we observe that these methods tend to fail choosing the most informative views when images are taken with non-uniform spacing.
	As shown in \reffig{fig:delaunay}, we may not be able to reconstruct the whole novel view for the target position (yellow dot) depending on scene geometry when choosing the right three views (blue dots) in case of large occlusions. Choosing based on proximity can force the network to extrapolate data from the source views while---ideally---we would want to network to interpolate the data in order to achieve smooth transitions and handle occlusions.
	
	Therefore, we propose to choose the "working set" based on the Delaunay triangulation of the view positions. This ensures both a better coverage of the nearby view space and allows for an easy way to compute weightings for the views by using barycentric coordinates. Since we construct the triangulation in 2D while the camera positions are in 3D, we first need to determine which view configuration is present in the scene.
	%	 and our cameras live in 3D space, the first step is to determine which view configuration is present for the scene.
	We distinguish between two types: hemisphere sampling in which the views are placed in the upper hemisphere around the scene and fronto-parallel sampling where they are mostly planar in front of the scene.
	
	For the case of hemisphere sampling, we stereographically project the camera positions onto the 2D plane where we perform the triangulation and then lift the result back to 3D. For fronto-parallel sampling, we orthographically project the camera positions onto the common plane defined by all views. 
	
	After triangulation, the closest triangle to the target view position is selected and the corresponding images form the working set. These three images are also assigned a weight $b_i$ which corresponds to the barycentric coordinate of the target view w.r.t. the triangle. \reffig{fig:delaunay} (right) shows selected view positions with our approach on the previous example.
	
	Finally, we use a shared U-Net model to extract feature maps $\mathbf{F}_i\in\mathbb{R}^{H_i\times W_i\times d}$ for each image $\mathbf{I}_i$ in the working set. 
	
	To be noted that we only consider a working set of three images. A generalization to more images would require changing the view selection strategy and is left for future work.

\bgroup
\def\W{60pt}
\def\H{10pt}
\def\Sep{25pt}
\def\ShiftLevel{20pt}
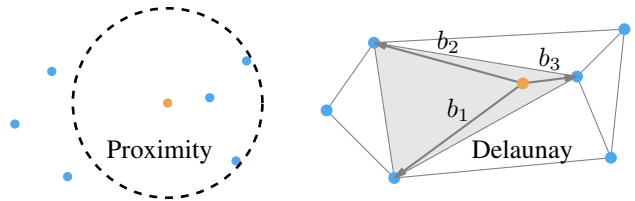
\begin{figure}[]
	\centering
	\begin{tikzpicture}
	
	%%%POINETNET
	\node[] at (0,0) {
		\begin{tikzpicture}[scale=0.7, >={Stealth[inset=0pt,length=6pt,angle'=28,round]} ]
		
%		\node[] (dummy) at (-4, 0){};

		%points on the left side
		\node[coordinate] (pt1) at (0, 0){}; %middle
		\node[coordinate] (pt2) at (1, -1){};  %lower
		\node[coordinate] (pt3) at (0.7, 1){}; %upper
		%points on the right side  
		\node[coordinate] (pt4) at (3+0.7, 0.5){}; %middle
		\node[coordinate] (pt5) at (3.5+0.7, -0.7){}; %lower
		\node[coordinate] (pt6) at (3.7+0.7, 1.2){}; %upper
		\fill[ph-blue!70] (pt1) circle (2.5pt);
		\fill[ph-blue!70] (pt2) circle (2.5pt);
		\fill[ph-blue!70] (pt3) circle (2.5pt);
		\fill[ph-blue!70] (pt4) circle (2.5pt);
		\fill[ph-blue!70] (pt5) circle (2.5pt);
		\fill[ph-blue!70] (pt6) circle (2.5pt);
		
		%draw lines
%		%triangles on the left side
%		\draw[thin,gray] (pt1)--(pt2);
%		\draw[thin,gray] (pt1)--(pt3);
%		\draw[thin,gray] (pt2)--(pt3);
%		%triangles on the right side
%		\draw[thin,gray] (pt4)--(pt5);
%		\draw[thin,gray] (pt4)--(pt6);
%		\draw[thin,gray] (pt5)--(pt6);
%		%lines in between
%		\draw[thin,gray] (pt2)--(pt5);
%		\draw[thin,gray] (pt3)--(pt6);
%		\draw[thin,gray] (pt2)--(pt4);
%		\draw[thin,gray] (pt3)--(pt4);
		
		%make node in the middle for the novel view
		\node[coordinate] (n) at (2.4+0.5, 0.4){}; %upper
		\fill[ph-orange!70] (n) circle (2.5pt);
		
		%circle 
%		\filldraw (n) circle (3pt);
		\draw [dashed, line width=1pt] (n) circle(1.8cm);

		\end{tikzpicture}
		
	};
	%with delaunay
	\node[] at (4.5,0) {
		\begin{tikzpicture}[scale=0.9,>={Stealth[inset=0pt,length=6pt,angle'=28,round]} ]
		
%		\node[] (dummy) at (-4, 0){};
		
		%points on the left side
		\node[coordinate] (pt1) at (0, 0){}; %middle
		\node[coordinate] (pt2) at (1, -1){};  %lower
		\node[coordinate] (pt3) at (0.7, 1){}; %upper
		%points on the right side  
		\node[coordinate] (pt4) at (3+0.7, 0.5){}; %middle
		\node[coordinate] (pt5) at (3.5+0.7, -0.7){}; %lower
		\node[coordinate] (pt6) at (3.7+0.7, 1.2){}; %upper
			
		%fill the triangle 
		\fill[fill=gray!20] (pt2.center)--(pt3.center)--(pt4.center);

		%draw lines
		%triangles on the left side
		\draw[thin,gray] (pt1)--(pt2);
		\draw[thin,gray] (pt1)--(pt3);
		\draw[thin,gray] (pt2)--(pt3);
		%triangles on the right side
		\draw[thin,gray] (pt4)--(pt5);
		\draw[thin,gray] (pt4)--(pt6);
		\draw[thin,gray] (pt5)--(pt6);
		%lines in between
		\draw[thin,gray] (pt2)--(pt5);
		\draw[thin,gray] (pt3)--(pt6);
		\draw[thin,gray] (pt2)--(pt4);
		\draw[thin,gray] (pt3)--(pt4);

		\fill[ph-blue!70] (pt1) circle (2.5pt);
		\fill[ph-blue!70] (pt2) circle (2.5pt);
		\fill[ph-blue!70] (pt3) circle (2.5pt);
		\fill[ph-blue!70] (pt4) circle (2.5pt);
		\fill[ph-blue!70] (pt5) circle (2.5pt);
		\fill[ph-blue!70] (pt6) circle (2.5pt);
		
		\node[coordinate] (n) at (2.4+0.5, 0.4){}; %upper
		
		%draw arrows towards the selected ones	
		\draw [thick, draw=gray, ->] (n) -- (pt2) node[midway, above, black] {$b_1$};
		\draw [thick, draw=gray, ->] (n) -- (pt3) node[midway, above, black] {$b_2$};
		\draw [thick, draw=gray, ->] (n) -- (pt4) node[midway, above, black] {$b_3$};

		%make node in the middle for the novel view
		
		\fill[ph-orange!70] (n) circle (2.5pt);
		
		%circle 
%		\filldraw (n) circle (3pt);
%		\draw [dashed, line width=1pt] (n) circle(1.7cm);

		\end{tikzpicture}
		
	};
	
%	
%	%%ResNetblock
%	\node[] at (4.5,-3.7) {
%	
%		
%	};
	
		\node[text width=1cm] at (0.1,-.65) {Proximity};
		\node[text width=1cm] at (4.95,-.65) {Delaunay};

\end{tikzpicture}
\vspace*{-2ex}
\caption{Given a novel view (orange) we need to choose a working set from all the views in the dataset (blue). Proximity-based view selection can lead to significant occlusion as the working set only views the scene from the right. Our Delaunay-based approach selects the views that belong to the closest triangle, ensuring scene viewing from different sides.} \label{fig:delaunay}
\end{figure}
\egroup
	
		\subsection{Geometry}

	\bgroup
	\def\Img{./images/multires2/multires2.png} 
	\newlength{\WMR}
	\newlength{\HMR}
	\settowidth{\WMR}{\includegraphics{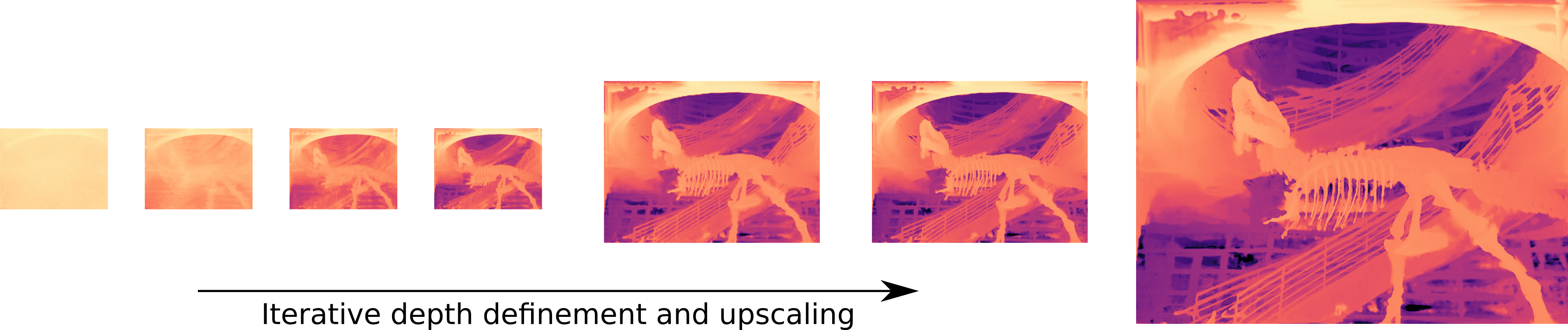}}
	\settoheight{\HMR}{\includegraphics{\Img}}
	\begin{figure*}
		\begin{center}
			\includegraphics[trim=.0\WMR{} 0.0\HMR{} 0.0\WMR{} 0.0\HMR{},clip, width=\textwidth]{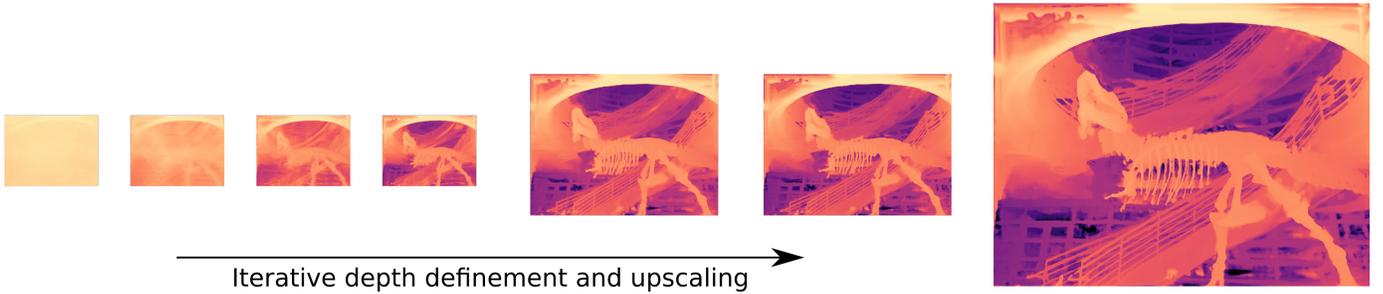}
		\end{center}
		\vspace*{-3ex}
		\caption{Depth estimation for the novel view starts on a coarse scale and is initialized to a constant value close to the camera. The network iteratively refines the current estimate using input view features and upscales the depth until the full resolution is reached.}
		\label{fig:multires}
	\end{figure*}
	\egroup

	To recover the scene geometry, we shoot rays from each pixel of the target view and find the intersection of the ray with the scene surface.
	\ac{nerf}-like models accomplish this by densely sampling the ray at predefined intervals. Hence, most samples lay in empty space, slowing down processing. In contrast, we draw inspiration from \ac{srn}~\cite{srn} and propose to use a differentiable sphere tracer which predicts for each sample on the ray a jump towards the next sample. This effectively enables the network to learn and adapt the step-size, which greatly improves the sample efficiency and speed.
	
	We parametrize each ray from the target view as follows:
	\begin{equation}
	r(t) = \mathbf{o} + t \mathbf{d},
	\end{equation}
	where $t$ is the distance along the ray, $\mathbf{o}$ is the origin of the camera in world coordinates, and $\mathbf{d}$ is the normalized direction of the ray. 	
	We initialize $t$ to be a small value such that ray marching starts close to the camera. 
	At each ray marching step, the position of the sample $\mathbf{x}=r(t)$ is obtained in world coordinates. The ray sample is projected into each source view $\mathbf{I}_i$ from where local features $\mathbf{f}_i\in\mathbb{R}^d$ are extracted using bilinear interpolation.
	The local features $\mathbf{f}_i$ from the source images are aggregated into a final feature by computing their weighted mean $\boldsymbol{\mu}$ and variance $\mathbf{v}$ using the barycentric weights.
	
	The aggregated features are also concatenated with the positional encoding of the ray samples which helps the network to recover high-frequency depth. Hence, the aggregated feature for each ray is defined as:
	\begin{equation}
	\mathbf{g}= \left[  \boldsymbol{\mu}, \mathbf{v}, \gamma(\mathbf{x})    \right],
	\end{equation}
	
	where $\gamma(.)$ is a positional encoding mapping the position into a higher-dimensional space~\cite{nerf}.
	
	Before computing the jump towards the next sample, an important consideration is that a single point sample does not contain sufficient information for an accurate jump prediction. Many real-world scenarios have objects with poor texture or ambiguous depth. If each ray sample independently predicts its own jumps, the final depth map will end up being noisy. We argue here that having knowledge of how the neighbouring rays behave is crucial for resolving ambiguities. 
	%	Therefore, we add a series of $3\times3$ convolutions between the features of the neighbouring rays of the target view in order to better constrain their features.
	Therefore, we add a series of $3\times3$ convolutions after the feature aggregation step in order to better constrain the features of each ray. To be noted that the per-pixel embeddings from U-Net capture only information from one particular view while convolving on the ray features also reasons about the multi-view features from the working set of images.
	
	Finally, the ray features are passed through an LSTM that predicts the displacement $\delta$ along the ray which is used to update our depth $t_{i+1}=t_{i}+\delta$.
	This process is iterated a fixed number of times (we use 18 in our experiments) and the final ray sample is considered to be on the surface of the object.
	This is in stark contrast to \ac{nerf}-like models which require samples in the order of hundreds.
	
	Since time consumption increases with the number of ray marching iterations and the number of rays we traverse, we propose to alleviate this problem by employing a coarse-to-fine scheme. 
	Instead of creating rays for each pixel of the target view of size $H\times W$, we first ray march from a downsampled version at quarter resolution. After several ray marching steps, the computed depth map is upsampled bilinearly to half resolution and the ray marching continues. This process, shown in~\reffig{fig:multires}, iterates until the final full resolution is reached. We observe that this scheme works well since locally-close pixels tend to march together and therefore their depth can be recovered by marching them as a whole. 
	We use three levels of hierarchical depth, each with 10, 5, and 3 ray march steps, respectively.
	
	\def\Img{./images/nerf_synthetic/nerf_synth.png} 
%	\begin{wrapfigure}{R}{0.5\textwidth}
	\begin{figure}
%		\begin{center}
		\centering
		\includegraphics[width=0.45\textwidth]{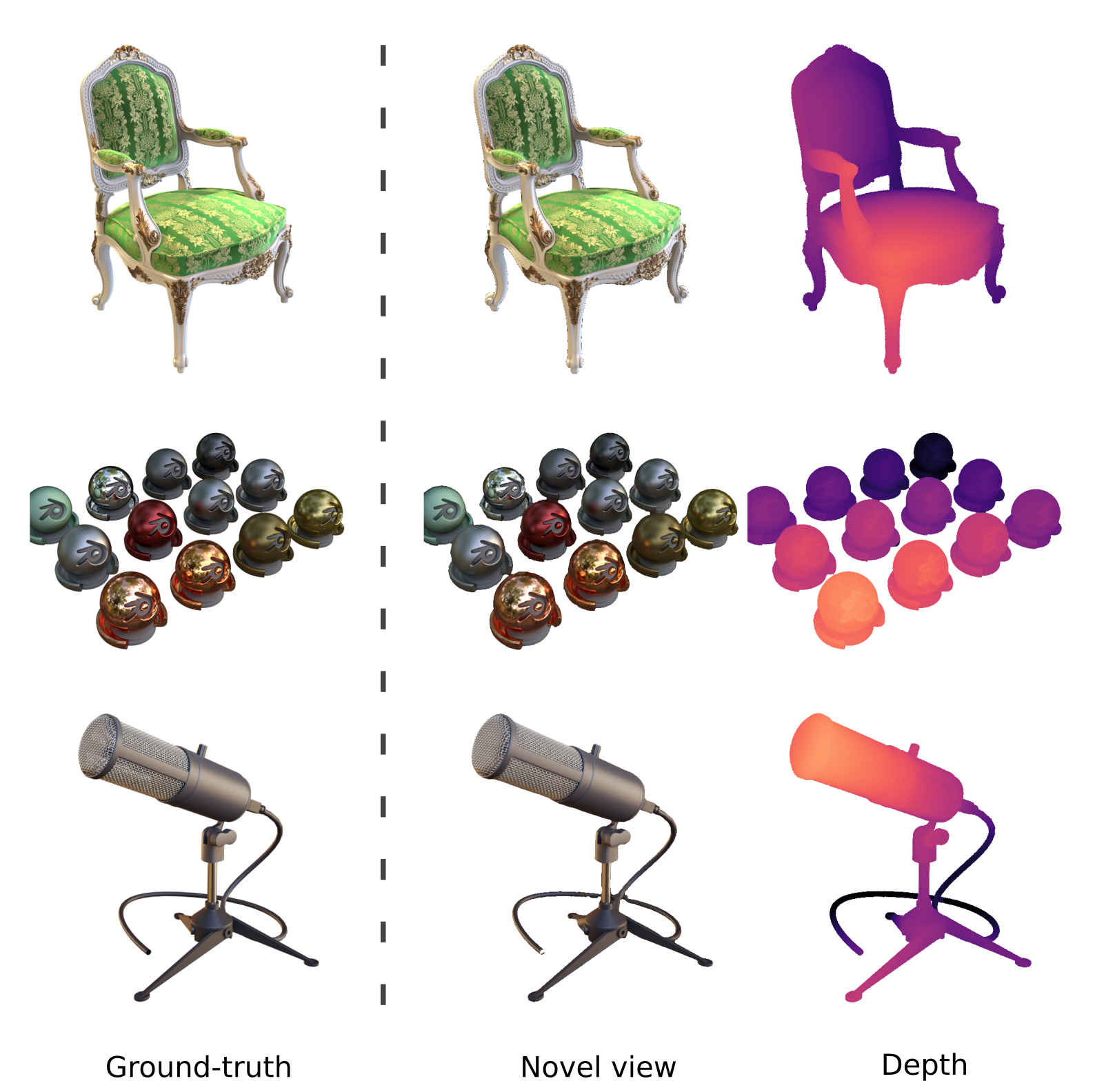}
		\caption{Results from the Realistic Synthetic dataset~\cite{nerf}. Our network captures both high-frequency detail and view-dependent effects like specular reflections.}
		\label{fig:nerf_synth}
%	\end{wrapfigure}
	\end{figure}

\subsection{Color} 

	We obtain per-pixel color by projecting the final ray-marched surface into the three input views and bilinearly sampling both color $\mathbf{c}_i$ and local features $\mathbf{f}_i$ which are concatenated together in $\mathbf{k}_i=\left[ \mathbf{c}_i, \mathbf{f}_i  \right]$. Instead of aggregating $\mathbf{k}_i$ using the barycentric weights, we observed that it is beneficial to allow the network to predict the weights. Therefore, similar to IBRNet~\cite{ibrnet}, we first compute $\boldsymbol{\mu}$ and $\mathbf{v}$ using the barycentric weights in order to capture global information. Afterwards, we concatenate these aggregated features with each per-frame feature vector $\mathbf{k}_i$.  Each concatenated feature is fed into a small MLP to integrate both local  and global information and predict multi-view aware feature $\mathbf{k}^\prime_i$ and blending weights $w_i\in[0, 1]$. We pool $\mathbf{k}^\prime_i$ into mean and variance by using the weights $w_i$ and map the resulting vector to RGB color using another MLP. We denote the final RGB image with $\mathbf{\tilde{I}}$.

	The color loss is computed as the $\ell_1$-loss between the recovered RGB and the ground-truth color. This loss implicitly biases the geometry to lie on the true scene surface since the correct depth produces consistent input view features and the color prediction becomes possible.
	This allows the network to learn unsupervised depth and be applicable to datasets with only RGB images.
	
	The reader should further note that we output a full RGB map in one pass of our network. In contrast, \ac{nerf}-like methods output a limited number of pixels at a time since their ray-marching step is more expensive and therefore requires to run the network multiple times to complete the full image. This allows our method to use more complex losses like perceptual losses which need to operate on the full image.

\subsection{Loss with Confidence Estimation}

	Apart from predicting a correct novel view, it is also valuable to predict a confidence for each pixel. This allows to reason about possible occlusions or regions which are outside the frustum of the input views. In essence, the network can hallucinate detail when needed but it should be aware of this hallucination. The confidence is not used directly in our network but it is a useful output for downstream tasks like depth fusion.

	In order to predict a confidence map, we draw inspiration from the work of Wagner \etal~\cite{wagner2019interpretable} which attempt to recover fine-grained explanations from classification networks. The input to their classification network is a pixel-wise blend between the image and a zero image. The loss function attempts to set as many pixels as possible to zero without affecting the classification accuracy. Hence, non-zero pixels are the ones that the network deems important for classification.
	
	In our approach, we choose a similar scheme by defining our loss as a blend between the predicted $\mathbf{\tilde{I}}$ and the ground-truth image $\mathbf{I}$. The blend uses the confidence map $\mathbf{Q}$ which encourages it to be as close as possible to $\num{1}$ such that most of the pixels are chosen from the predicted image. Then our image loss with confidence estimation is defined as:
	
	\begin{equation}
	L= \norm{ \mathbf{I} -  \left(  \mathbf{\tilde{I}}\cdot \mathbf{Q} +  \mathbf{I}\cdot (1-\mathbf{Q})\right)  }_1 + \lambda\norm{1-\mathbf{Q}}_2.
	\end{equation}
	
	\begin{figure}[]
		\centering
		\includegraphics[width=1.0\linewidth]{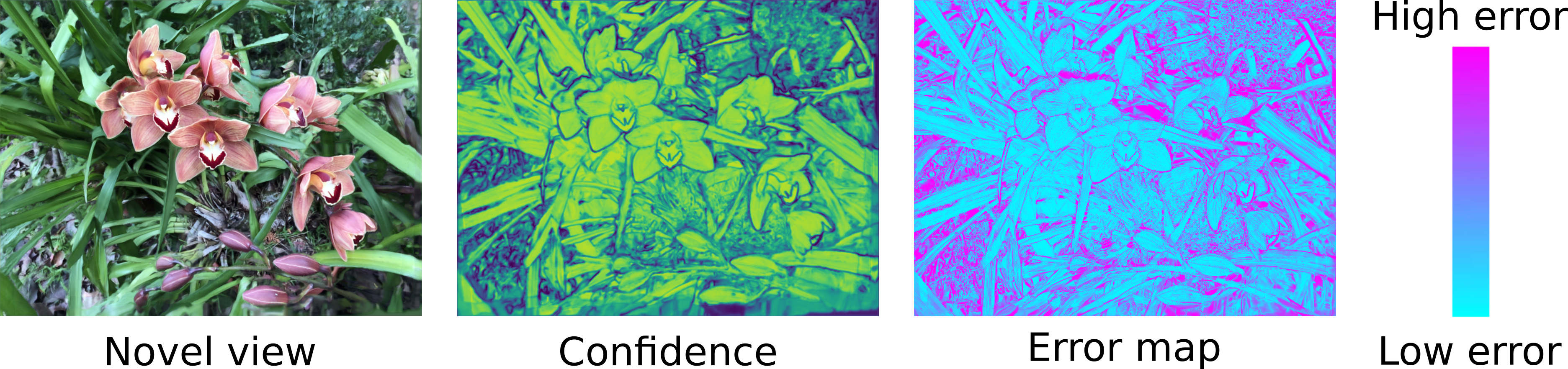}
		\vspace*{-2ex}
		\caption{The confidence predicted by the network is low near depth discontinuities where occlusion occur and therefore errors are likely.} \label{fig:confidence_explanation}
	\end{figure}

\section{Results}

\subsection{Datasets}
\begin{figure}[]
		\centering
		\includegraphics[width=1.0\linewidth]{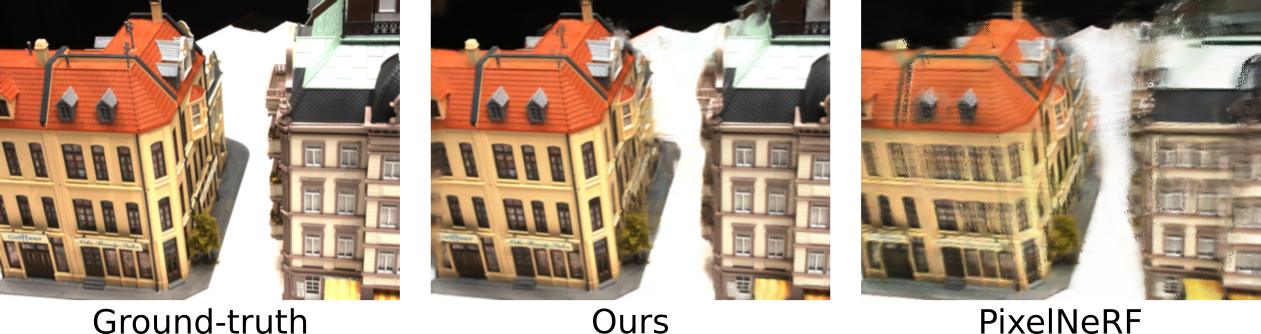}
		%		\vspace*{-2ex}
		\caption{ Comparison of novel views on the test set of DTU. The model didn't use any per-scene optimization and was trained only on the training set of DTU showing that it can generalize to novel views and novel objects.  } \label{fig:dtu_comparison}
	\end{figure}

	We evaluate our method on three datasets. 	
	DTU~\cite{dtu_dataset} contains real images of various objects and is targeted towards evaluation of MVS methods. We use the train and test splits as defined by PixelNeRF~\cite{pixelnerf}: 88 scenes for training and 15 for testing at a resolution of $400\times300$. We use this dataset to test the generalization capabilities of our method. The objects in the test set are different from the ones in the training set, so if the network is able to recover novel views of these novel objects, we can conclude that it learned a general reconstruction method. Results of the generalization to novel objects and novel views can be seen in~\reffig{fig:dtu_comparison}.
	
	Realistic Synthetic $360^\circ$~\cite{nerf} contains synthetic images of objects from the upper hemisphere. The dataset contains eight scenes with images at $800\times800$ resolution. The objects exhibit several view-dependent effects like specular reflections which must be captured correctly by the network for properly rendering the target view. Results of our network's prediction on this dataset can be seen in~\reffig{fig:nerf_synth}.
	
	Real Forward-Facing~\cite{mildenhall19} consists of real images of large scenes scanned with a camera in a forward-facing manner. The dataset contains eight scenes with image size of $1008\times756$. We use the train and test split as defined by~\cite{ibrnet}: every 8th image is selected for testing.
	
	\begin{figure*}[]
		\centering
		\includegraphics[width=1.0\textwidth]{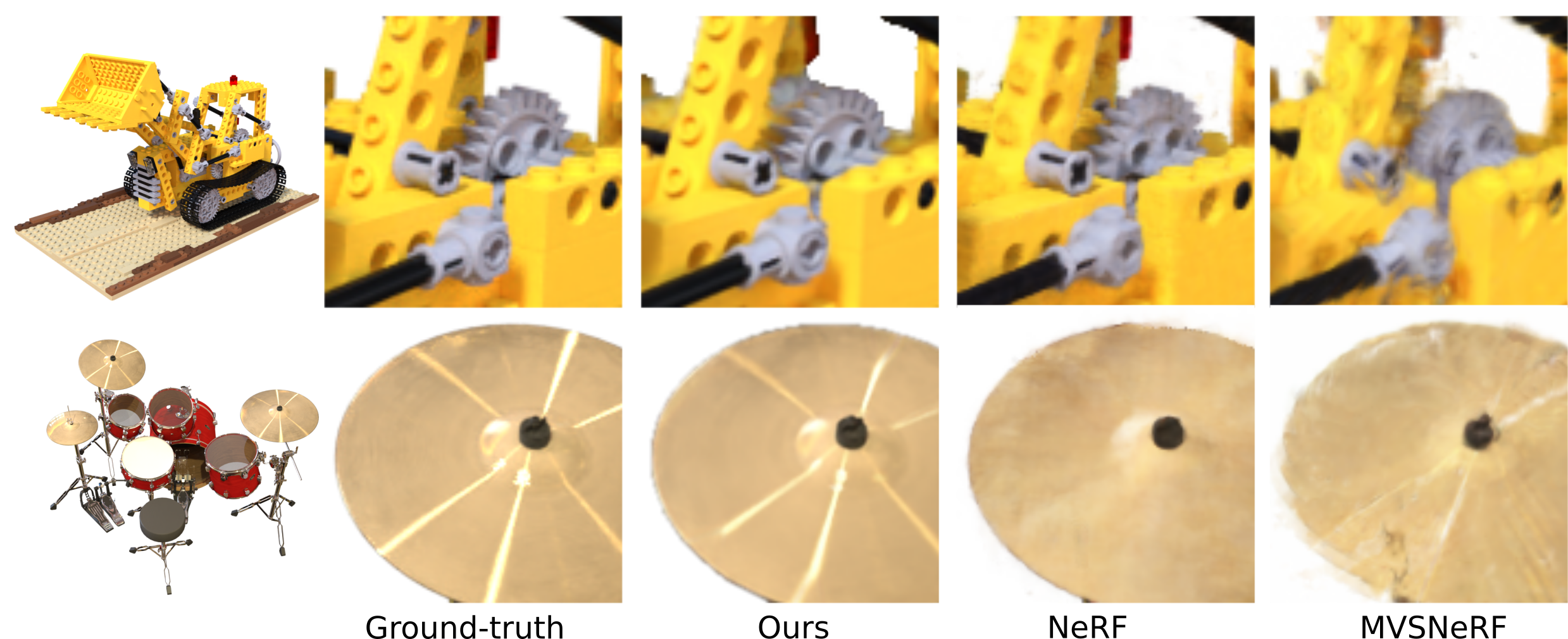}
		%		\vspace*{-2ex}
		\caption{ Comparison of test-set views of the synthetic dataset. Our method can recover sharp detail together with view-dependent effects. However, because our method is image-based, it struggles with occlusions and thin objects like the drum stands. } \label{fig:nerf_comparison}
	\end{figure*}	

\subsection{Evaluation}

	We train our method on the three datasets and distinguish between with and without per-scene optimization. 
	In the case of no scene optimization, we train a generalizable model on the DTU dataset~\cite{dtu_dataset} and evaluate on the synthetic~\cite{nerf}, a real dataset~\cite{mildenhall19}, and the novel scenes from DTU. \reftab{tab:results} shows that our network generalizes to the novel views despite the drastic change in scale and object types. We also train our model with per-scene images similar to \ac{nerf} and show that it performs comparable to other generalizable models like MVSNeRF~\cite{mvsnerf} while being significantly faster.
	
	In~\reffig{fig:nerf_comparison} we compare our per-scene optimized model with the other baselines. We observe that our model can recover more detail especially in highly specular areas. However, our method also exhibits more errors near occlusion boundaries. This is to be expected as our method is imaged-based and therefore areas which are occluded in all source images cannot be reliably reconstructed.
%		but has more error in occluded regions, unlike NeRF-like approaches which don't suffer from this. 

	%The results from baselines methods are obtained from their official implementations or from the reported evaluations in the respective articles.

	%%%%% small results table

% \newcommand{\tablespace}{\,\,\,}
\newcommand{\tablespace}{\,\,\,\,}
\newcommand{\halftablespace}{\,}
\setlength{\tabcolsep}{4pt}
\begin{table*}[t]
\centering

\caption{Comparison of different methods on multiple real and synthetic datasets. Results with ${}^\dagger$ correspond to NeRF model trained for 9.5h as evaluated by MVSNeRF~\cite{mvsnerf}.}

\resizebox{\textwidth}{!}{%

%%%attemot 3 
\begin{tabular}{l|c|ccc|ccc|ccc}
%realisitc, dtu, llff
\multirow{2}{*}{Method} & \multirow{2}{*}{Setting} & \multicolumn{3}{c|}{DTU~\cite{dtu_dataset}} & \multicolumn{3}{c|}{Realistic Synthetic $360^\circ$~\cite{nerf}} & \multicolumn{3}{c}{Real Forward-Facing~\cite{mildenhall19}} \\
 && PSNR$\uparrow$ & SSIM$\uparrow$ & LPIPS$\downarrow$ & PSNR$\uparrow$ & SSIM$\uparrow$ & LPIPS$\downarrow$ & PSNR$\uparrow$ & SSIM$\uparrow$ & LPIPS$\downarrow$ \\
\hline
% $\dagger$ &DeepVoxels~\cite{deepvoxels} & $30.55$ & $0.97\phantom{0}$ & - & - & - & - & - & - & - \\
%SRN~\cite{srn}  & \multirow{3}{*}{\shortstack{No per-scene \\ optimization}} & $33.20$ & $0.963$ & $0.073$ & $22.26$ & $0.846$ & $0.170$ & $22.84$ & $0.668$ & $0.378$ \\

pixelNeRF~\cite{pixelnerf}  & \multirow{4}{*}{\shortstack{No per-scene \\ optimization}} & $24.14$ & $0.887$ & $ 0.224$ & $4.36$ & $0.46$ & $0.44$ & $11.266$ & $0.388$ & $0.757$ \\
IBRNet~\cite{ibrnet}  && $25.84$ & $0.902$ & $0.213$ & $19.43$ & $0.841$ & $0.231$ & $16.70$ & $0.566$ & $0.498$\\
MVSNeRF~\cite{mvsnerf}  && $25.17$ & $\mathbf{0.911}$ & $0.185$ & $\mathbf{22.67}$ & $\mathbf{0.90}$ & $\mathbf{0.21}$ & $17.56$ & $\mathbf{0.691}$ & $0.381$ \\ 
Ours  && $\mathbf{26.376}$ & $0.896$ & $\mathbf{0.184}$ & $20.070$ & $0.687$ & $0.242$ & $\mathbf{18.909}$ & $0.643$ & $\mathbf{0.372}$ \\ 

\midrule

%NeRF$_{9.5h}$~\citep{•}cite{nerf}    &  \multirow{3}{*}{\shortstack{Per-scene \\ optimization}} & $23.70$ & $0.893$ & $0.247$ & $26.95$ & $0.939$ & $0.136$ & $23.67$ & $0.820$ & $0.339$ \\
NeRF~\cite{nerf}    &  \multirow{3}{*}{\shortstack{Per-scene \\ optimization}} & $23.70^\dagger$ & $0.893^\dagger$ & $0.247^\dagger$ & $\mathbf{31.01}$ & $0.947$ & $ 0.081$ & $\mathbf{26.50}$ & $0.811$ & $0.250$ \\
MVSNeRF~\cite{mvsnerf}    && $\mathbf{29.30}$ & $\mathbf{0.959}$ & $\mathbf{0.101}$ & $27.21$ & $ 0.945$ & $0.227$ & $26.25$ & $\mathbf{0.907}$ & $\mathbf{0.139}$ \\
Ours    && $28.093$ & $0.913$ & $0.165$ & $28.425$ & $\mathbf{0.952}$ & $\mathbf{0.070}$ & $25.206$ & $0.803$ & $0.218$  \\

 \hline
\end{tabular}
}

\label{tab:results}  
%\caption{Comparison of different methods on multiple real and synthetic datasets. Results with ${}^\dagger$ correspond to NeRF model trained for 9.5h as evaluated by MVSNeRF~\cite{mvsnerf}.}
\end{table*}
\setlength{\tabcolsep}{1.4pt}

	\subsection{Performance} 
	
	Previous methods are unable to process the full image at once due to the high computational demand per ray and thus need to run several times with different ray batches to complete the image. We set the ray batches to the maximum size that fits in the memory of an NVIDIA GeForce RTX 3090 and measure the time for rendering a novel view with a resolution of $800\times800$ pixels. \reftab{tab:speed} shows that our method renders the full image in one forward pass and requires significantly less time than all previous approaches.
	
%	\begin{table}
%		\centering
%		\begin{tabularx}{\linewidth}{ C C }
%			\resizebox{0.35\columnwidth}{!}{%
%				\begin{tabular}{l|c @{\hskip 0.2in} c}
%					%realisitc, dtu, llff
%					Method & Time & Rays  \\
%					\hline
%					
%					\textcolor{red}{MVSNeRF}~\cite{mvsnerf}  & \textcolor{red}{\SI{5.2}{\second}} & \textcolor{red}{\SI{110}{\kilo\nothing}} \\
%					NeRF~\cite{nerf}  & \SI{6.4}{\second} & \SI{120}{\kilo\nothing} \\
%					IBRNet~\cite{ibrnet}  & \SI{31}{\second} & \SI{8}{\kilo\nothing} \\
%					pixelNeRF~\cite{pixelnerf}  & \SI{164}{\second}  & \SI{300}{\kilo\nothing} \\
%					\midrule
%					%Ours & \boldmath{\SI{0.16}{\second}}  & $\bm{full}$(\SI{640}{\kilo\nothing}) \\
%					Ours & $\mathbf{0.16~\text{\bfseries s}}$ & $\mathbf{full(640~\text{\bfseries k})}$\\
%					\hline
%					
%				\end{tabular}
%			}
%			\caption{Computation time and maximum rays per batch for a $800\times 800$ image.}\label{tab:speed}
%			&
%			\resizebox{0.5\columnwidth}{!}{%
%				\begin{tabular}{l|ccc}
%					Setting & PSNR$\uparrow$ & SSIM$\uparrow$ & LPIPS$\downarrow$  \\
%					\hline
%					No position encoding & $25.692$ & $0.899$ & $0.082$ \\
%					No Delaunay & $26.764$ & $0.919$ & $0.075$ \\
%					Fewer ray marches & $27.522$ & $0.933$ & $0.058$ \\
%					Only $1\times1$ convolutions & $26.224$ & $0.912$ & $0.075$ \\
%					\hline
%					Complete model & $\mathbf{27.918}$ & $\mathbf{0.937}$ & $\mathbf{0.056}$ \\
%					\hline
%					
%				\end{tabular}
%			}
%			\caption{Ablation study}\label{tab:ablation}
%		\end{tabularx}
%	\end{table}

	\begin{table}
		
		\centering
		
		\caption{Computation time and maximum rays per batch for a $800\times 800$ image.}\label{tab:speed}
		\resizebox{0.65\columnwidth}{!}{
		\begin{tabular}{l|c @{\hskip 0.2in} c}
			%realisitc, dtu, llff
			Method & Time & Rays  \\
			\hline
			
			MVSNeRF~\cite{mvsnerf}  & \SI{5.2}{\second} & \SI{110}{\kilo\nothing} \\
			NeRF~\cite{nerf}  & \SI{6.4}{\second} & \SI{120}{\kilo\nothing} \\
			IBRNet~\cite{ibrnet}  & \SI{31}{\second} & \SI{8}{\kilo\nothing} \\
			pixelNeRF~\cite{pixelnerf}  & \SI{164}{\second}  & \SI{300}{\kilo\nothing} \\
			\midrule
			%Ours & \boldmath{\SI{0.16}{\second}}  & $\bm{full}$(\SI{640}{\kilo\nothing}) \\
			Ours & $\mathbf{0.16~\text{\bfseries s}}$ & $\mathbf{full(640~\text{\bfseries k})}$\\
			\hline
			
		\end{tabular}
		}
	
%		\caption{Computation time and maximum rays per batch for a $800\times 800$ image.}\label{tab:speed}
	\end{table}

	\begin{table}
		\centering
		
		\caption{Ablation study}\label{tab:ablation}
		\resizebox{0.8\columnwidth}{!}{
		\begin{tabular}{l|ccc}
			Setting & PSNR$\uparrow$ & SSIM$\uparrow$ & LPIPS$\downarrow$  \\
			\hline
			No position encoding & $25.692$ & $0.899$ & $0.082$ \\
			No Delaunay & $26.764$ & $0.919$ & $0.075$ \\
			Fewer ray marches & $27.522$ & $0.933$ & $0.058$ \\
			Only $1\times1$ convolutions & $26.224$ & $0.912$ & $0.075$ \\
			\hline
			Complete model & $\mathbf{27.918}$ & $\mathbf{0.937}$ & $\mathbf{0.056}$ \\
			\hline
			
		\end{tabular}
		}
%		\caption{Ablation study}\label{tab:ablation}
	\end{table}

	\subsection{Ablation Study}
	
	We perform an ablation study of the different components of our network. 
	We train on the Lego scene from the synthetic dataset and observe how the network performance is affected.
	
	We first remove the positional encoding from the ray marching. This decreases the depth quality significantly as the network is unable to recover high-frequency details. 
	
	Disabling Delaunay for view selection and using a proximity-based method that takes the three closest frames as the working set shows also a slight decrease in performance.
	
	By default, we use three levels of depth refinement, each with 10, 5, and 3 ray march steps, respectively. We reduce the number of ray march steps to 5, 3, and 1 and observe a slight decrease in performance. The required number of ray marching steps is heavily dependent on the scene complexity with simple scene requiring less steps.
	
	Finally, we change the $3\times3$ convolutions in the ray marcher to $1\times1$ in order to simulate propagating each ray independently with no spatial awareness of the neighbouring rays, similar to other NVS methods. We observe a significant decrease in accuracy, as the rays can no longer leverage spatial information to resolve ambiguities. 
	
	\subsection{Implementation Details} 
		The feature extraction network is a U-Net model~\cite{unet} which outputs per-pixel a 64 dimensional vector. The features from the three images are aggregated and then passed through two convolutional layers of $3\times3$ which output 64 channels. Finally, the LSTM that predicts the sample jump has a hidden size of 32. The color estimation is implemented as an MLP with 3 layers and a hidden size of 64.
		%The network is optimized using Adam~\cite{kingma2014adam} with a learning rate of \num{1e-4}.
		We release the source code at \ttfamily{\url{https://github.com/AIS-Bonn/neural_mvs}}
		\normalfont{}

	\subsection{Limitations}
	\begin{figure}
		\centering
		\includegraphics[width=.9\linewidth]{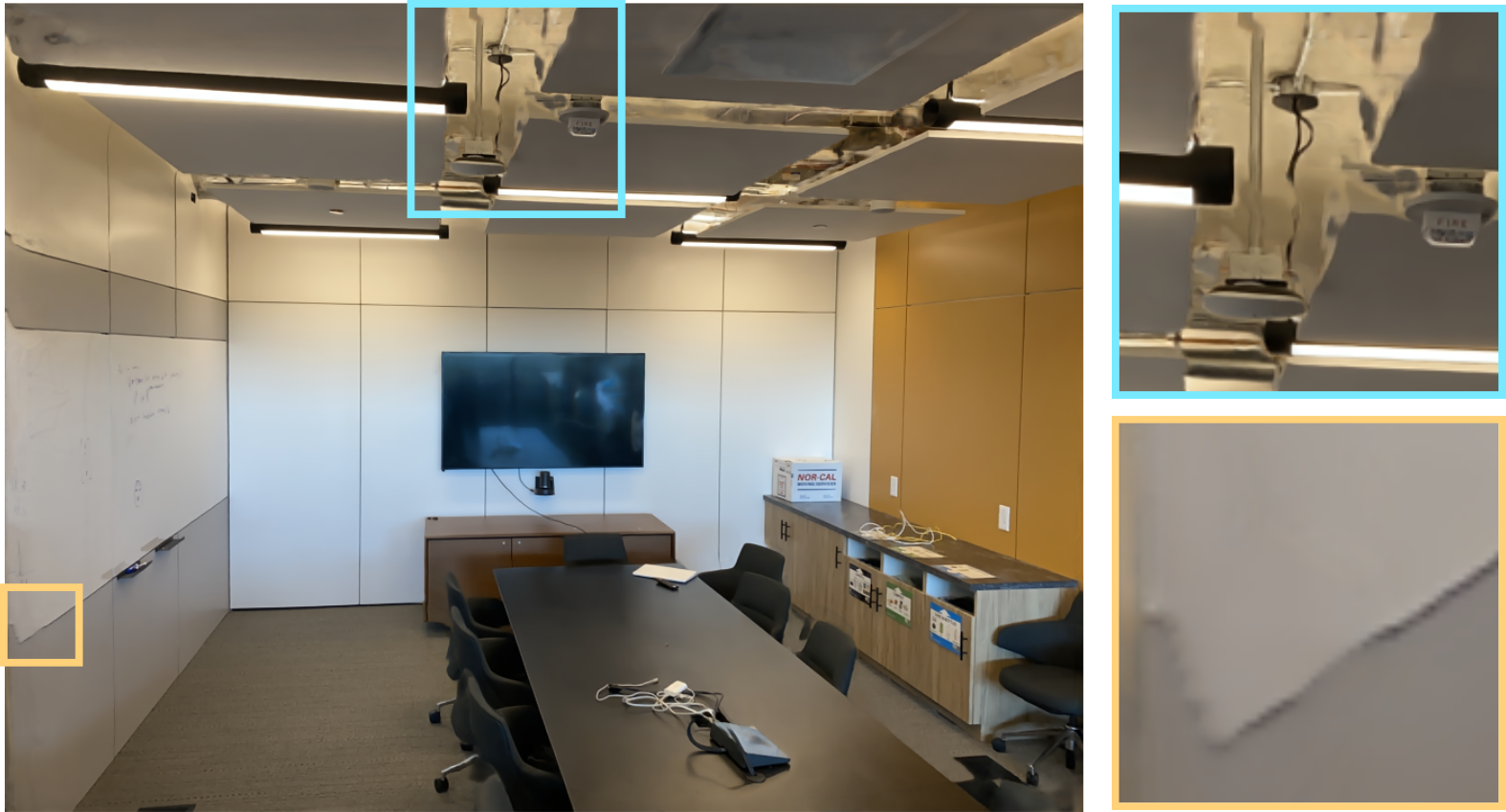}
		%		\vspace*{-2ex}
		%		\caption{ \textcolor{red}{Failure cases: Our method fails in the case of heavy occlusion as it's required to halucinate detail that is not present in any of the input views. This can be seen for example at the ceiling of the room in the front-facing dataset and at the edges of the image.} } \label{fig:failure}
		\caption{ Failure cases: Our method fails in the case of occlusion like the ceiling detail or the image borders. } \label{fig:failure}
	\end{figure}
	
		One limitation of our method is that it is based on ray-marching instead of volumetric rendering and therefore cannot model transparent objects. A switch to a front-to-back additive blending of radiance could alleviate this issue.

		Another limitation is that our method is image-based and therefore cannot recover detail in occluded regions as seen in~\reffig{fig:failure}.

		Finally, the depth can be ambiguous in the case of no texture since the network can recover correct color even if the depth is noisy. This could be alleviated by using more input views or with stronger priors.

\section{Conclusion}
	%	The use of coarse-to-fine hierarchy, the use of iterative depth refinement and view conditioning can faster network that can rival that can reconstruct 3D supervised only with images
	We proposed a network that jointly resolves scene geometry and novel view synthesis from multi-view datasets and is supervised only by image reconstruction loss. We represent the scene geometry as a distance function which we ray march using sphere tracing. Sphere tracing alleviates the memory constraints faced by other methods and allows us to render high resolution images in one forward pass and is thus much faster than previous methods. We further improve the speed by proposing a hierarchical depth refinement which estimates depth in a coarse-to-fine manner. 
	
	Finally, we show the generalization capabilities of our network by evaluating on datasets with different scale and object configurations for which we obtain competitive results but with significantly higher frame rates.

\bibliographystyle{IEEEtran}
\bibliography{egbib}

\end{document}